\documentclass[review]{elsarticle}

\usepackage{lineno}
\usepackage{hyperref}
\usepackage[justification=centering]{caption} 
\usepackage[export]{adjustbox}
\usepackage{subcaption}
\usepackage{amsmath}
\usepackage[linesnumbered,boxed]{algorithm2e}
\usepackage{booktabs}
\usepackage{multirow}
\usepackage{color,soul}  
\soulregister\cite7      
\soulregister\ref7		 
\modulolinenumbers[5]

\journal{Arxiv}

\begin{document}

\begin{frontmatter}

\title{Enhanced Discrete Particle Swarm Optimization Path Planning for UAV Vision-based Surface Inspection}

\author[label1]{Manh Duong Phung}
\ead{duongpm@vnu.edu.vn}
\author[label1]{Cong Hoang Quach}
\ead{hoangqc@vnu.edu.vn}
\address[label1]{Vietnam National University, 144 Xuan Thuy, Cau Giay, Hanoi, Vietnam}

\author[label2]{Tran Hiep Dinh}
\ead{tranhiep.dinh@uts.edu.au}
\author[label2]{Quang Ha\corref{cor1}}
\ead{quang.ha@uts.edu.au}
\address[label2]{University of Technology Sydney, 15 Broadway, Ultimo NSW 2007, Australia}
\cortext[cor1]{Corresponding author}

\begin{abstract}
In built infrastructure monitoring, an efficient path planning algorithm is essential for robotic inspection of large surfaces using computer vision. In this work, we first formulate the inspection path planning problem as an extended travelling salesman problem (TSP) in which both the coverage and obstacle avoidance were taken into account. An enhanced discrete particle swarm optimisation (DPSO) algorithm is then proposed to solve the TSP, with performance improvement by using deterministic initialisation, random mutation, and edge exchange. Finally, we take advantage of parallel computing to implement the DPSO in a GPU-based framework so that the computation time can be significantly reduced while keeping the hardware requirement unchanged. To show the effectiveness of the proposed algorithm, experimental results are included for datasets obtained from UAV inspection of an office building and a bridge.
\end{abstract}


\end{frontmatter}


\section{Introduction}
For robotics inspection of built infrastructure, computer vision can be used to detect most surface deficiencies such as cracking, spalling, rusting, distortion, misalignment, and excessive movements. Over the last decade, much research effort has been devoted to this theme with computer vision becoming an important component of modern Structural Health Monitoring (SHM) systems for built infrastructure such as rust detection of steel bridges \cite{Liao2016}, crack detection of concrete  bridges \cite{YeumD15, Li201483, Adhikari2014180}, or bridge condition assessment \cite{Zaurin2010}. In this regard, it is promising to integrate a computer vision system into mobile inspection robots, such as unmanned aerial vehicles (UAVs) \cite{Alex2016, Ellenberg2016} or ubiquitous robots \cite{Bock2015, Ha2011}, especially when dealing with large and hardly accessible structures like tunnels \cite{Montero201599}. For this purpose, an efficient inspection path planning (IPP) algorithm is therefore of crucial importance. 

In vision-based inspection path planning, it is required to find a trajectory that is informative enough to collect data from different views of a given structure so that the inspection robot can carry out the data acquisition of the region of interest. Depending on size of the inspecting region, the trajectory can be planned for multiple robots to coordinately conduct the data collection \cite{Zhang2014}. To be visibly processed at a later time, the data collected are often from a sensor of the time-of-flight (optical, sonar or radar) or passive optical (CCD camera) type. Since the computational time for IPP rapidly increases with the area of the region of interest, an IPP algorithm should meet the following criteria:

(i) capability of viewing/covering every surface of the region of interest via at least one planned viewpoint of the inspection sensor, 

(ii) obstacle avoidance for the robot,

(iii) generation of an "optimal" path under available conditions, and

(iv) effectiveness in terms of processing time (for online re-planning and large structure inspection).

Studies on IPP, in general, can be categorised into three groups, namely cell decomposition, sub-problem separation, and other methods. In cell decomposition, the target space is decomposed in sub-regions called cells. The cell shape can be trapezoidal, square, cubic, or customised depending on critical points of Morse functions, often with a uniform size \cite{Choset2001, Choset2005, Acar2002}. An exhaustive path connecting each cell is then computed for the coverage, using typically a heuristic algorithm such as wavefront \cite{Shivashankar2011} or spiral spanning tree \cite{Gabriely2002}. Methods based on cell decomposition yield good results in terms of coverage and obstacle avoidance. As the path generated, however, may not be optimal, it is worth seeking a better and more feasible alternative. In this context, the IPP separation approach tackling the non-deterministic polynomial time (NP)-hard problems can be divided into two, the art gallery problem that finds the smallest set of viewpoints to cover the whole gallery, and the travelling salesman problem (TSP) that finds the shortest path to visit a set of given cities \cite{Englot2013, Hollinger2013, Petr2013, Wang2011, Blaer2009, Saha2006}. Each problem can be solved separately using known methods such as the randomised, incremental algorithm for the art gallery problem \cite{Danner2000, Englot2016} and the chained Lin-Kernighan heuristics for the TSP \cite{Applegate2003}. Other approaches have focused on sampling the configuration space \cite{Papadopoulos2013}, using sub-modular objective function \cite{Hollinger2012}, or employing genetic algorithms \cite{Jimenez2007} but they often require constraining the robot to certain dynamic models or end with near-optimal solutions. The requirements remain not only a shorter path but also collision-free. 

In this paper, the IPP problem is addressed by first formulating it as an extended TSP. The enhanced discrete particle swarm optimisation (DPSO) is then employed to solve the IPP. Finally, parallel computing based on graphical processing units (GPU) is deployed to obtain the real-time performance. The contributions of our approach are three folds: (i) By formulating the IPP as an extended TSP, both the coverage and obstacle avoidance are simultaneously taken into account. In addition, constraints related to the kinematic and dynamic models of the robot are separated from the DPSO solution so that this solution can be applied to a broad range of robots. (ii) Three techniques including deterministic initialisation, random mutation, and edge exchange have been proposed to improve the accuracy of DPSO. (iii) Parallel computation has been implemented to significantly improve the time performance of DPSO. By utilising GPU, the parallel implementation does not add additional requirements to the hardware, i.e. the developed software can run on popular laptop computers.

The rest of this paper is structured as follows. Section~\ref{sect:formulation} introduces the steps to formulate the IPP as an extended TSP. Section~\ref{sect:dpso} presents the proposed DPSO and its deployment for solving the IPP. Section~\ref{sect:experiment} provides experimental results. Finally, a conclusion is drawn to end our paper.

\section{Problem formulation}
\label{sect:formulation}
Our ultimate goal is to design a path planning system for an UAV used for inspecting planar surfaces of largely built structures like buildings or bridges. The sensor used for the inspection is a CCD camera attached to a controllable gimbal. We suppose that the 3D model of the structure and the environment are known prior to planning, for example, by using laser scanners. Here, the IPP objective is to find the shortest path for the UAV's navigation and taking photos of the target surfaces so that the images captured can be later processed to detect potential defects or damages. We first consider the IPP as an extended TSP and then solve it using the developed DPSO. This section presents the computation of viewpoint selection and point-to-point pathfinding, which are fundamental to formulate the extended TSP problem.

\subsection{Viewpoint selection}
\begin{figure}
	\centering
	\includegraphics[scale=0.8]{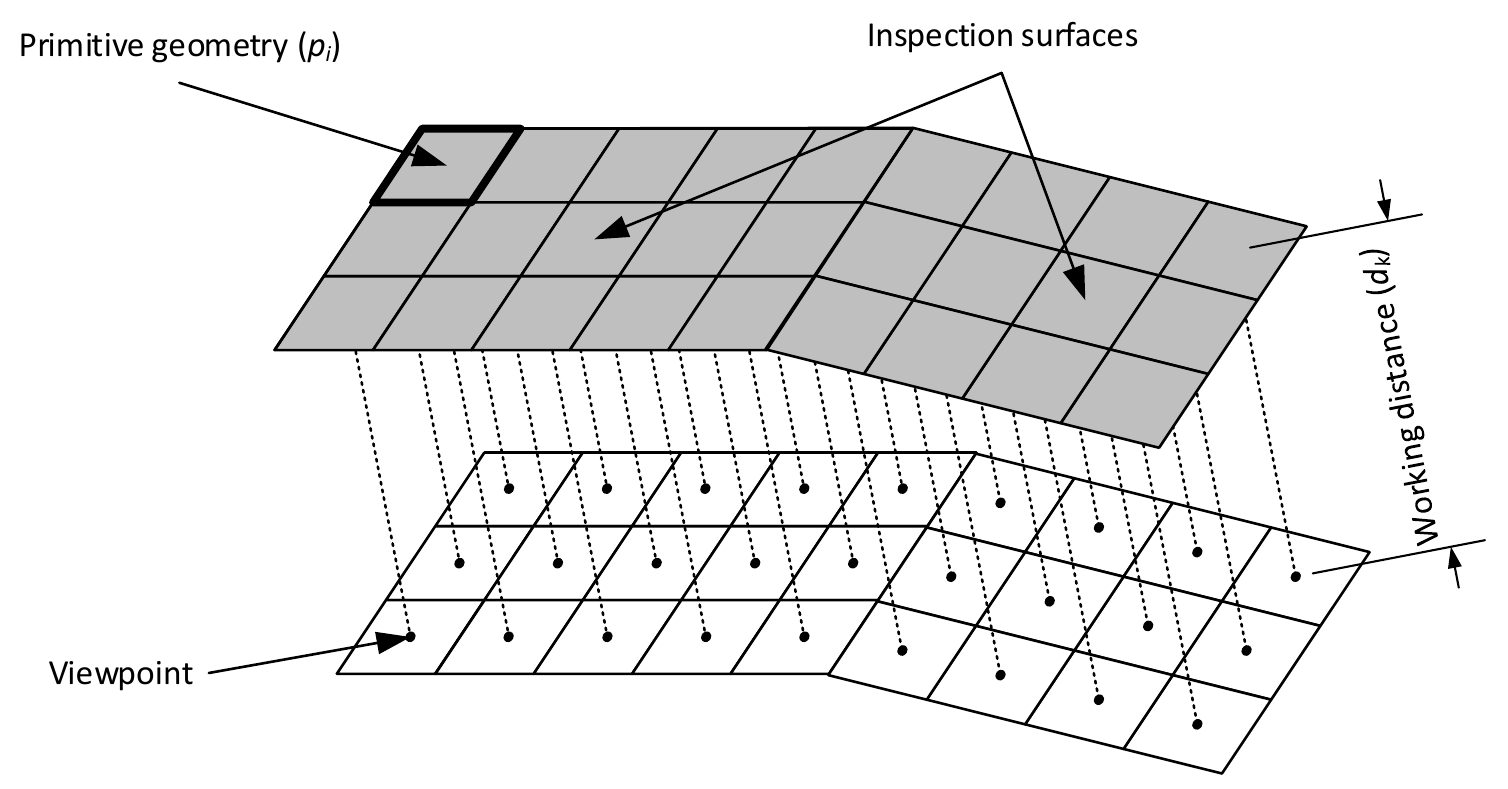}
	\caption{Generation of inspection viewpoints.}
	\label{fig:viewpoint}
\end{figure}

The viewpoint selection involves finding a set of camera configurations that together cover the whole surfaces of interest. Let $P$ be a finite set of geometric primitives $p_i$ comprising the surfaces to be covered. Each geometric primitive $p_i$ corresponds to a surface patch within the field of view of the camera. Let $C$ be the configuration space such that every feasible configuration $c_j \in C$ maps to a subset of $P$. Each configuration $c_i$ corresponds to a position $(x_i,y_i,z_i)$ and an orientation $(\varphi_i,\theta_i,\psi_i)$ of the camera. Given a finite set of configurations $C$, the viewpoint selection problem on one hand calls generally for the minimum number of configurations $c_i$ such that all elements $p_i \in P$ are covered. On the other hand, from image sticking and defect detection, the following requirements are added to the system:
(i) image capturing moment is when the camera is perpendicular to the inspected surface,
(ii) sufficiently high resolution to distinguish the smallest feature, $s_f$,  and 
(iii) overlapping of images to a percentage $o_p$ specified by the sticking algorithm. 
\begin{figure}[h!]
	\begin{subfigure}{0.5\textwidth}
		\centering
		\includegraphics[width=190 pt]{Figure/Camera1.pdf}
		\caption{}
	\end{subfigure}%
	\begin{subfigure}{0.5\textwidth}
		\centering
		\includegraphics[width=100	pt]{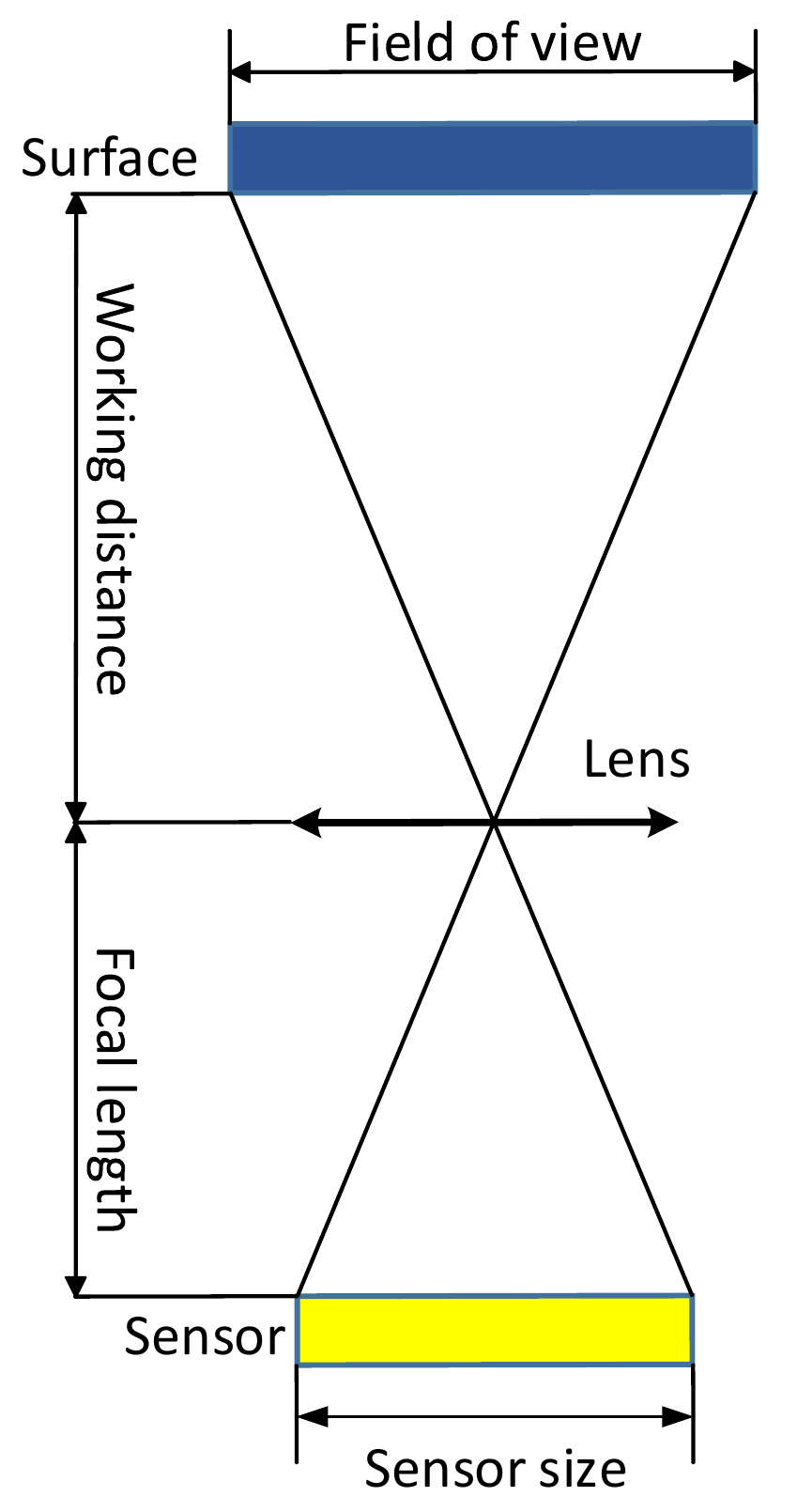}
		\caption{}
	\end{subfigure}
	\centering
	\caption{Camera for inspection: (a) Camera setup in the field; (b) Relation between parameters of the camera and the field.}
	\label{fig:cameraSetup}
\end{figure}
It turns out that those requirements simplify our selection problem. The perpendicular requirement confines the camera orientation to the normal of the inspected surface. The resolution requirement suggests the computation of the field of view of the camera as:

\begin{equation}
a_{fov} = \frac{1}{2} r_c s_f,
\label{eq:fov}
\end{equation}
where $r_c$ is the camera resolution (see Fig.~\ref{fig:cameraSetup}). Taken the overlapping percentage into account, the geometric primitive $p_i$ is then:

\begin{equation}
p_i = (1 - o_p)  a_{fov}.
\label{eq:premitive}
\end{equation}
The working distance from the camera to the surface can also be computed as:
\begin{equation}
d_k = \frac{a_{fov} f}{s_s},
\label{eq:cameraDistance}
\end{equation}
where $f$ and $s_s$ are respectively the focal length and sensor size of the camera. From (\ref{eq:premitive}) and (\ref{eq:cameraDistance}), it is possible to determine configurations $c_i$ to cover the set of primitives $P$, as illustrated in Fig.~\ref{fig:viewpoint}. Specifically, for each surface $P_k \subset P$, a grid with the cell size of $p_i$ is first established to cover $P_k$. A working surface $P_k^*$, parallel to $P_k$ and distant $d_k$ from $P_k$, is then created. Projecting the center of each cell of $P_k$ to $P_k^*$ gives the position component of viewpoint $c_i$. The normal of $P_k$ defines the orientation component of $c_i$, which is supposed to be fully controlled by the inspecting UAV so that it can be omitted in our computation.

\subsection{Point-to-point pathfinding}
\label{sect:path}
Given the viewpoints, the shortest, obstacle-free path between every pair of them need be found to form a graph for later processing. Without loss of generality, different motion planning approaches such as roadmap, decoupling, potential field and mathematical programming can be used here depending on the UAV model and dynamic constraints \cite{Goerzen2009,Dadkhah2012}. In this work, the hierarchical decoupled approach is employed in which open- and closed-loop controllers operating at a variety of rate are linked together from top to bottom \cite{Goerzen2009,Scherer2008,Jung2013}. Since the majority of UAVs currently in production often already equipped with an inner-loop tracking controller and a waypoint following system, this approach can be simplified to a discrete search that produces a set of waypoints connecting two viewpoints while avoiding obstacles. For this, the workspace is first divided into a grid of voxels. Each voxel has the free or occupied status corresponding to the presence or absence of an object in that voxel. In order to consider the UAV as a particle moving without collision between voxels, all the free voxels in a sphere of a radius equal to the largest dimension of the UAV are marked as occupied. Thus, the A* algorithm \cite{Hart1968} can be used to find the shortest path between viewpoints. In each step, the cost to move from one voxel to another surrounding neighbour is computed as:
\begin{equation}
L(\alpha,\beta,\gamma)=a_1 \alpha^2+ a_2 \beta^2+a_3 \gamma^2, 
\end{equation}
where coordinates $\alpha,\beta,\gamma \in \{-1,0,1\}$  indicate the position of neighbor, and coefficients $a_1$, $a_2$ and $a_3$ assign a particular weight to each direction. The total cost to move from a voxel $p$ to the viewpoint $g$ at step $n$ is given by:

\begin{equation}
f(p) = \sum\limits_{k=1}^n L_k + \lVert p - g \lVert^2,
\label{eq:cost}
\end{equation}
where $L_k$ is the motion cost at step $k$. 

\subsection{Modelling the IPP as a TSP}
For given viewpoints and paths between them, a graph can be built to model the IPP as an extended TSP. We define each viewpoint as a node, $i$, and the path between two viewpoints as an edge, $e_{ij}$. The length, $l_{ij}$, of edge $e_{ij}$ is the cost to travel from node $i$ to node $j$ determined by (\ref{eq:cost}). If the path between node $i$ and node $j$ is blocked due to obstacles, a virtual path between them is defined and a very large cost is assigned for the path. Denoting the set of all nodes by $V$ and the set of all edges by $E$, we restrict motion of the UAV to the graph $G = (V, E)$. The IPP task is now to find a tour, with a minimum cost, that visits each node (viewpoint) exactly once, including the way back to the initial node. Let $T$ be the set of these nodes.

By associating a binary variable 
\begin{equation}
\lambda_{ij} = \left\{\begin{array}{ll}
1 \quad \text{if edge $e_{ij} \in E$ is in tour}\\
0 \quad \mathrm{otherwise}
\end{array}\right. 
\end{equation}
with each edge in the graph, the IPP is then formulated as follows:

\begin{equation}
\label{eq:min}
\text{min} {\displaystyle \sum_{e_{ij} \in E} l_{ij} \lambda_{ij}}
\end{equation}
\begin{equation}
\label{eq:twoedge}
\text{subject to} \quad {\displaystyle \sum_{j \in V, \, i \neq j } \lambda_{ij}} = 2 \quad \forall i \in V
\end{equation}      	
\begin{equation}
\label{eq:nosubtour}
{\displaystyle \sum_{i,j \in T, \, i \neq j} \lambda_{ij}} \leq  \left\vert{T}\right\vert - 1 \quad \forall T \subset V, T \ne \emptyset 
\end{equation}
\begin{equation}
\label{eq:binary}
\quad \lambda_{ij} \in \{ 0, 1 \},
\end{equation}
where $\left\vert{T}\right\vert$ is the number of nodes in the tour.
The objective function in (\ref{eq:min}) defines the shortest tour. The constraint in (\ref{eq:twoedge}) implies each node in the graph has exactly one incoming edge and one outgoing edge, i.e., the tour passes through each node once, while condition (\ref{eq:nosubtour}) ensures no sub-tours, i.e., the tour returns to the original node after visiting all other nodes. 

\section{Enhanced Discrete Particle Swarm Optimization for Inspection Path Planning}
\label{sect:dpso}

Particle swarm optimization (PSO), inspired by social behavior of bird flocking or fish schooling, is a population-based stochastic technique designed for solving optimization problems \cite{Kennedy2001}. In PSO, a finite set of particles is generated, each particle seeks the global optimum by moving and evolving through generations. Initially, each particle is assigned to a random position and velocity. It then moves by updating its best previous position, $P_k$, and the best position of the swarm, $G_k$. Let $x_k$ and $v_k$ be respectively the position and velocity of a particle at generation $k$. The position and velocity of that particle in the next generation is given by:
\begin{equation}
\label{eq:velocity}
v_{k+1} \leftarrow w.v_k + \varphi_1r_1.(P_k - x_k) + \varphi_2r_2.(G_k - x_k)
\end{equation}
\begin{equation}
\label{eq:position}
x_{k+1} \leftarrow x_k + v_{k+1} ,
\end{equation}
where $w$ is the inertial coefficient, $\varphi_1$ is the cognitive coefficient, $\varphi_2$ is the social coefficient, and $r_1$, $r_2$ are random samples of a uniform distribution in the range [0,1]. Equations (\ref{eq:velocity}) and (\ref{eq:position}) imply that the motion of a given particle is a compromise between three possible choices including following its own way, moving toward its best previous position, or toward the swarm's best position. The ratio between choices is determined by the coefficients $w$, $\varphi_1$, and $\varphi_2$. 

\subsection{DPSO approach to the IPP}
Since the IPP defined in (\ref{eq:min}) -- (\ref{eq:binary}) is a discrete optimization problem, enhanced algorithms for discrete particle optimization (DPSO) will be developed for our problem, motivated by \cite{clerc2004}. For this, let us begin with an outline of our approach to solve the IPP problem using  DPSO with improvements in initialization, mutation, edge exchange and parallel implementation. 

First, let define the position of particles as sequences of $N+1$ nodes, all distinct, except that the last node must be equal to the first one: 
\begin{equation}
\label{eq:sequence}
x = (n_1, n_2, ..., n_N, n_{N+1}), n_i \in V, n_1 = n_{N+1},
\end{equation}
where $N$ is the number of nodes, $N = |V|$. Since each sequence is a feasible tour satisfying (\ref{eq:twoedge}) and (\ref{eq:nosubtour}), to minimise the objective function (\ref{eq:min}) according to (\ref{eq:velocity}) and (\ref{eq:position}), we need to define the velocity and numerical operators for the particles' motion.

From (\ref{eq:position}), it can be seen that a new position of a particle can be evolved from the position of its current generation via the velocity operator, considered here as a list of node transpositions: 
\begin{equation}
\label{eq:sequence}
v = ((n_{i,1}, n_{j,1}), (n_{i,2}, n_{j,2}),..., (n_{i,\lVert v \lVert},n_{j,\lVert v \lVert})),
\end{equation}
where $n_i,n_j \in V$ and $\lVert v \lVert$ is the length of the transposition list.

In DPSO, particle velocities and positions are updated by using the following operations:
\begin{itemize}
\item The $addition$ between a position $x$ and a velocity $v$ is found by applying the first transposition of $v$ to $x$, then the second one to the result, etc. For example, with $x=(1,4,2,3,5,1)$ and $v = ((1,2),(2,3))$, by applying the first transposition of $v$ to $x$ and keeping in mind the equality between the first and last nodes, we obtain (2,4,1,3,5,2). Then applying the second transposition of $v$ to that result gives (3,4,1,2,5,3), which is the final result of $x + v$. 

\item The $subtraction$ between a position $x_2$ and a position $x_1$ is defined as the velocity $v$, i.e., $x_2 - x_1=v$, such that by applying $v$ to $x_1$ we obtain back $x_2$.

\item The $addition$ between a velocity $v_1$ and a velocity $v_2$ is defined as a new velocity, $v_1 \oplus v_2=v,$ which contains the transpositions of $v_1$ followed by the transpositions of $v_2$. 

\item The $multiplication$ between a real coefficient $c$ with a velocity $v$ is a new velocity, $c.v$, defined as follows:
	\begin{itemize}
		\item For $c=0,~ c.v = \emptyset$.
		\item For $0 < c \le 1,~ c.v = ((n_{i,1}, n_{j,1}), (n_{i,2}, n_{j,2}),..., (n_{i,c\lVert v \lVert},n_{j,c\lVert v \lVert}))$.
		\item For $c < 0$ and $c > 1$, we omit these cases since they do not occur in our DPSO.
	\end{itemize}
\end{itemize}

\subsection{Augmentations to the DPSO}
In order to speed up the convergence and avoid being stuck in the local minimum, we propose to enhance optimisation performance of the DPSO as follows.
\subsubsection{Deterministic initialization}
The swarm in DPSO, having no prior knowledge of the searching space, is initialized with its particles at random positions. This initialization works well for a relatively small search space. 

\begin{figure}
	\centering
	\includegraphics[scale=0.5]{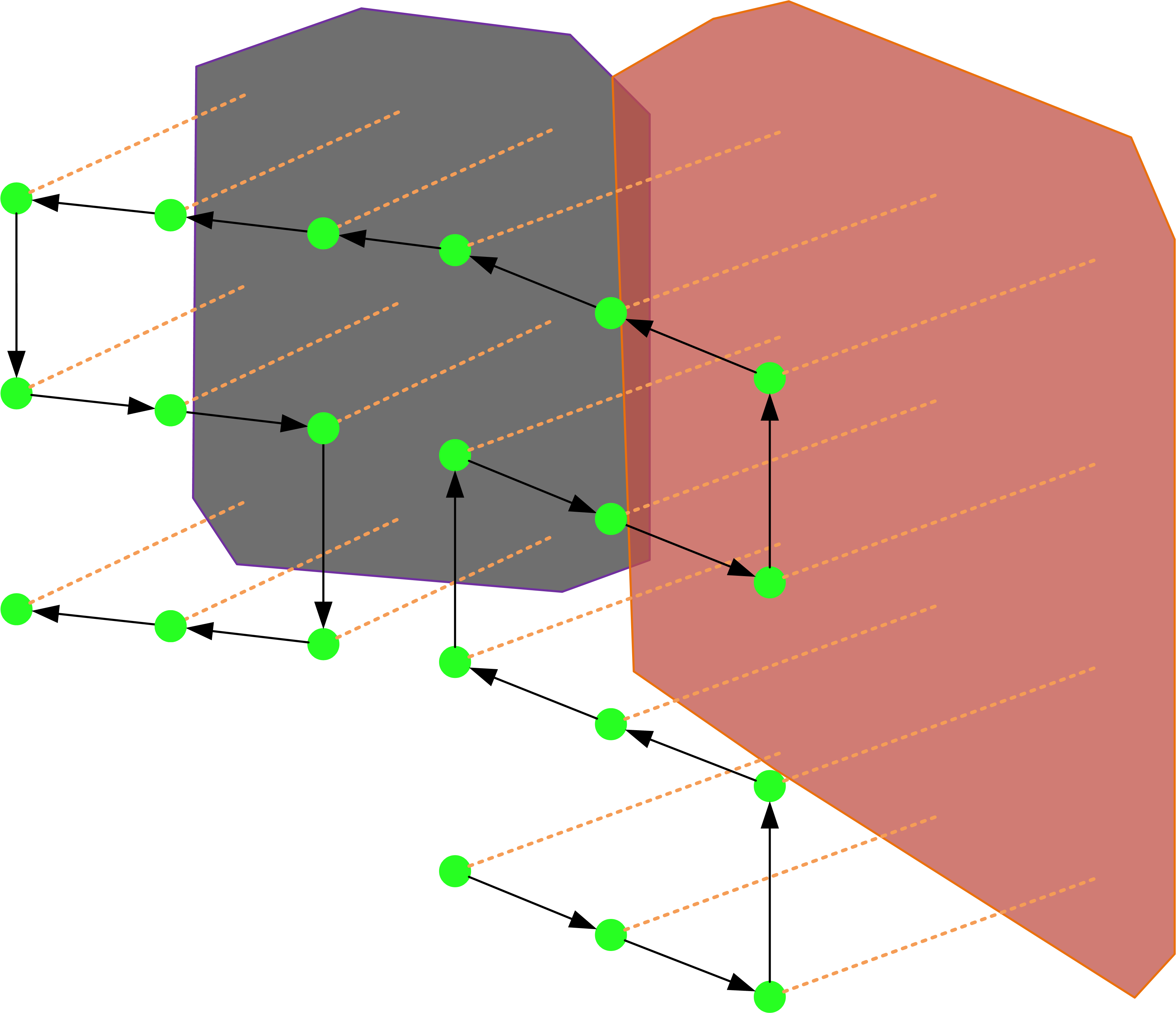}
	\caption{Initialization of particle using back-and-forth path.}
	\label{fig:initialization}
\end{figure}

For large structure, the searching result depends, to a great extent, on the initial positions of the particles. Therefore, in order to increase the probability of reaching the global optimum, we propose to exploit features of viewpoints to generate several seeding particles to facilitate the evolution of the swarm in the search space. In our application, viewpoints are generated based on a grid decomposition. Consequently, a back-and-forth tour would generate a near-optimal path, as shown in Fig.~\ref{fig:initialization}, if no obstacles occur. From this observation, positions are deterministically assigned for several particles during the initialization process. 

\subsubsection{Random mutation}
Similar to other evolutionary optimisation techniques such as the genetic algorithm or ant colony system, the PSO performs both exploration and exploitation of the search space. Initially, particles are far from each other so they explore different regions in the search space. After evolving through generations, the swarm converges and starts to make more exploitation. At this stage, distances between particles will gradually reduce to the size termed "swarm collapse" \cite{Kennedy2001}, whereby many particles will become almost identical. 

In order to avoid the collapse situation and keep the balance between exploration and exploitation, random mutations for particles are employed. After every $i$ generations, identical particles are filtered. The remaining are then sorted according to their cost values. Finally, only one-third of the smallest particles are kept for the next generation. All others are disturbed, each in different and randomly-chosen dimensions. 

\begin{figure}
	\centering
	\includegraphics[scale=0.95]{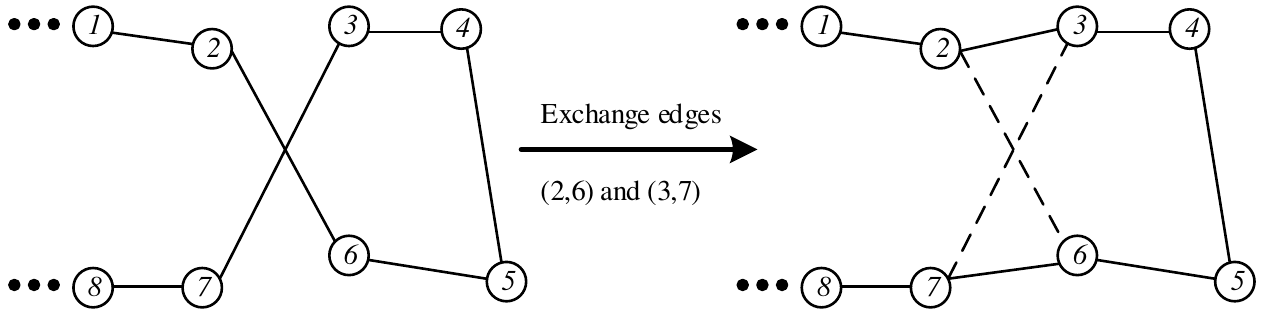}
	\caption{DPSO augmentation using edge exchange.}
	\label{fig:edge}
\end{figure}

\subsubsection{Edge exchange}  The enhancement is based on the geometric feature for which crossing edges can be exchanged to result in a shorter tour. Here, as 3D cross checking may be difficult, a complete search similarly to the 2-opt algorithm is employed to compare each valid combination of the swapping mechanism for edges \cite{Gutin2007}. In this search, every possible exchange of edges is evaluated and the one with the most improvement is chosen. Figure \ref{fig:edge} illustrates the case when an edge exchange between (2,6) and (3,7) to  shorten the tour. Since this augmentation is computational demanding, it should be used only when the random mutation does not make any difference.

\begin{figure}
	\centering
	\includegraphics[scale=0.8]{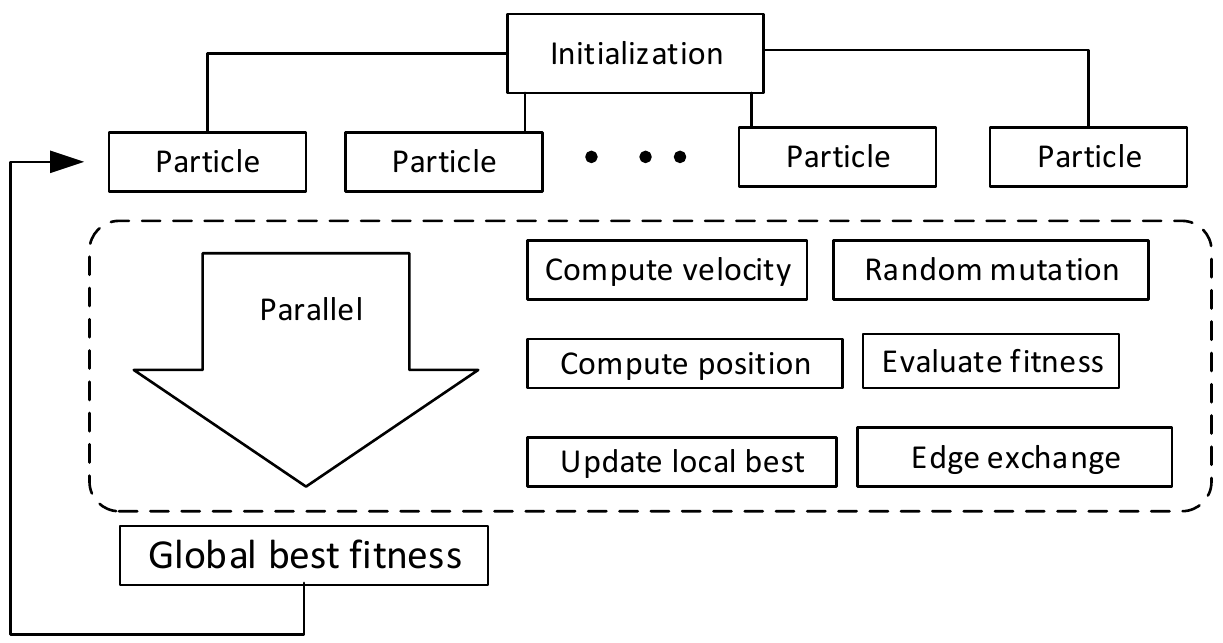}
	\caption{Parallel implementation of the DPSO on GPU.}
	\label{fig:parallel}
\end{figure}

\begin{figure}[h!]
	{\fontsize{08}{08}\selectfont
		\begin{algorithm}[H]
			\SetAlgoLined
			\tcc{Host:}
			Load $w$, $\varphi_1$, $\varphi_2$, $swarm\_size$ to global memory\;
			Load $graph$ of shortest paths and travelling costs to global memory\;
			Copy initalized $particles$ from CPU to global memory; \tcc*[f]{see Fig.\ref{fig:pseudocode} line 1-8} \\
			Set $threads\_per\_block$ = $swarm\_size$\;
			Call $kernels$ to evolve each particle in a seperate thread\;
			\tcc{Device:}
			\textbf{Kernel} $move\_particle$(*$particles$)\{\\
				\quad Get *$particle$ corresponding to $thread\_id$\;
				\quad Update $position$ and $fitness$ of $particle$; \tcc*[f]{see Fig.\ref{fig:pseudocode} line 10-17} \\
				\quad Update $global\_best$ to global memory\ and synchronize threads\;
			\}\\
			\textbf{Kernel} $random\_mutation$(*$particles$)\{\\
				\quad Sort $particles$ using \textbf{thrust} library\;
				\quad Randomize $2/3$ worst particles\;
			\quad Update $global\_best$ to global memory\ and synchronize threads\;
			\}\\
			\textbf{Kernel} $edge\_exchange$(*$graph$)\{\\
				\ForEach {$i < (number\_of\_nodes - 2)$} {
					\ForEach {$ i < j < (number\_of\_nodes - 2)$} {
						Swap $nodes(i,j)$ and evaluate $fitness$\;
					}
				}
			\quad Update $global\_best$ to global memory\ and synchronize threads\;
			\}\\
		\end{algorithm}
	}
	\caption{Pseudo code for parallel computation of DPSO on GPU.}
	\label{fig:GPUpseudocode}
\end{figure}

\subsubsection{Parallel implementation on GPU} Owing to the rapidly increased performance  with thousands of cores, a graphics processing unit (GPU) can outperform the traditional CPUs for problems that are suitable for processed by SIMD (single instruction multiple data). As our optimisation algorithms are also a SIMD-based, we can take this advantage to implement in parallel the proposed DPSO in GPUs to reduce computation time. 

The diagram and pseudo code for parallel implementation are shown in Fig.~\ref{fig:parallel} and Fig.~\ref{fig:GPUpseudocode} respectively. After initialization, parameters of a particle such as the velocity, position, and fitness are computed in parallel, each particle in a different thread. At the end of each generation, the results are saved to the global memory to update these particle parameters and then a new parallel computation round starts.

In UAVs, parallel programs can be implemented in recent onboard computers having good GPU capability and low power consumption such as Jetson TK1 with 192 CUDA Cores (Kepler GPU), 5 W \cite{Yash2015}. The board can be configured as either the main or supplemental board communicated with other components via standard communications protocols like MAVLink. However, if the battery power is highly limited as in some micro UAVs, an alternative solution is to stream the sensory data to the ground control station (GCS) and utilise the GPU of a laptop to conduct the path planning. The result is then uploaded to the UAV via GCS for planning/re-planning and navigation.

\subsection{Enhanced DPSO Pseudo Code}
For vision-based inspection, to take into account obstacle avoidance of the UAV, a selected combination of random and deterministic initialization for each particle in the swarm is performed on a CPU while its evolutions, including computation of updated particles' velocity and position, random mutation and edge exchange, are implemented in parallel on a GPU.

By making use of all advantages of the enhanced DPSO algorithm, the pseudo code for our proposed algorithm incorporating the above-mentioned augmentations is shown in Fig. \ref{fig:pseudocode}.

\begin{figure}[h!]
	{\fontsize{08}{08}\selectfont
		\begin{algorithm}[H]
			\SetAlgoLined
			\tcc{------------------------ Computation on CPU -------------------------}
			\tcc{Initialization:}
			Set swarm parameter $w$, $\varphi_1$, $\varphi_2$, $swarm\_size$\; 
			\ForEach {particle in swarm} {
				\quad	Initialize particle's position with 10\% specific and 90\% random\;
				\quad	Compute $fitness$ value of each particle\;
				\quad	Set $local\_best$ value of each particle to itself\; 
				\quad	Set velocity of each particle to zero\; 
			}
			Set $global\_best$ to the best fit particle\;
			\tcc{------------------------ Computation on GPU -------------------------}
			\tcc{Evolutions:}
			\Repeat {$max\_generation$ \textnormal{not reached} and \\ \qquad \space \space $global\_best$ \textnormal{not remaining unchanged for a pre-specified number of generations}} {
				\ForEach {particle in swarm} { 
					Compute new velocity; \tcc*[f]{using Eq.\ref{eq:velocity}} \\
					Compute new position; \tcc*[f]{using Eq.\ref{eq:position}} \\
					Update $fitness$ of new position\;
					\If{\textnormal{new} $fitness$ $<$ $local\_best$}{
						$local\_best$ =  \textnormal{new} $fitness$\;
					}
				}
				\If (\tcc*[f]{Random mutation}) {$current\_generation$ \textnormal{reaches} collapsed\_cycle} 				{ 
					Sort all particles by $fitness$\; 
					Randomize $2/3$ worst particles\;
				} 
				Find the particle with the best $fitness$ and update $global\_best$\;
				\If (\tcc*[f]{Edge exchange}) {$global\_best$ \textnormal{not improved}}{
					\ForEach {particle in swarm} {
						Swap each pair of nodes and evaluate $fitness$\;
						Choose the swap with best fit\; 
					}
				}
			}
		\end{algorithm}
	}
	\caption{Pseudo code of the enhanced DPSO algorithm.}
	\label{fig:pseudocode}
\end{figure}

\section{Experimental results}
\label{sect:experiment}
\begin{figure}[h!]
	
		\begin{subfigure}{0.5\textwidth}
		\centering
		\includegraphics[width=170 pt]{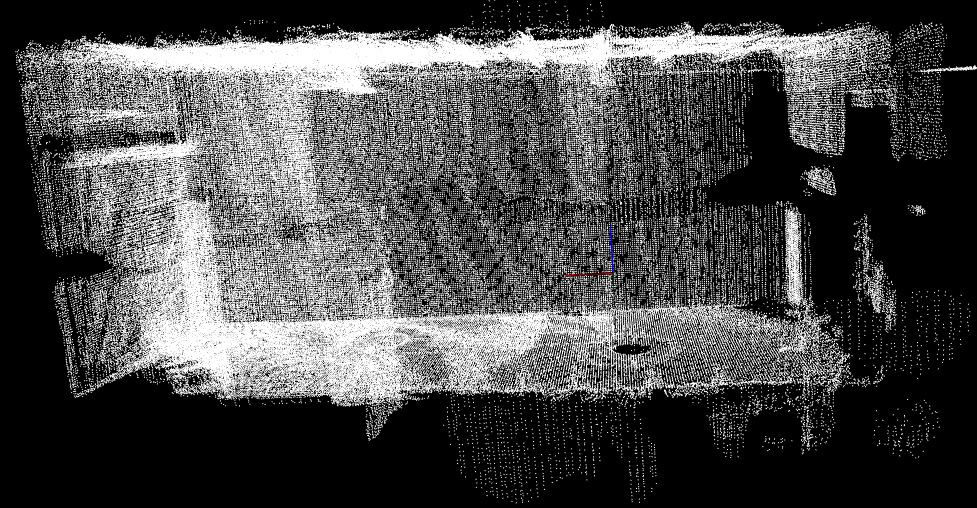}
		\caption{}
		\label{fig:office_a}
	\end{subfigure}%
	\begin{subfigure}{0.5\textwidth}
		\centering
		\includegraphics[width=170 pt]{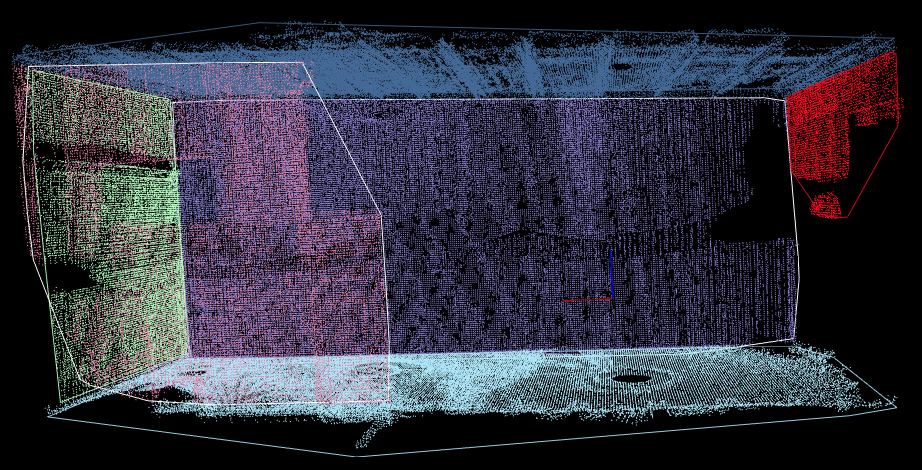}
		\caption{}
		\label{fig:office_b}
	\end{subfigure}
	\begin{subfigure}{0.5\textwidth}
		\centering
		\includegraphics[width=170 pt]{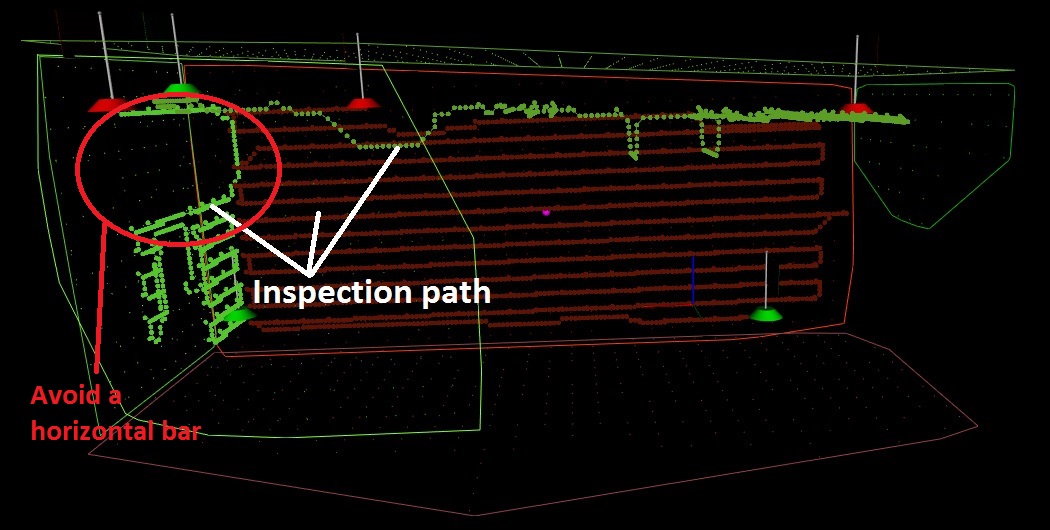}
		\caption{}
		\label{fig:office_c}
	\end{subfigure}%
	\begin{subfigure}{0.5\textwidth}
		\centering
		\includegraphics[width=170 pt]{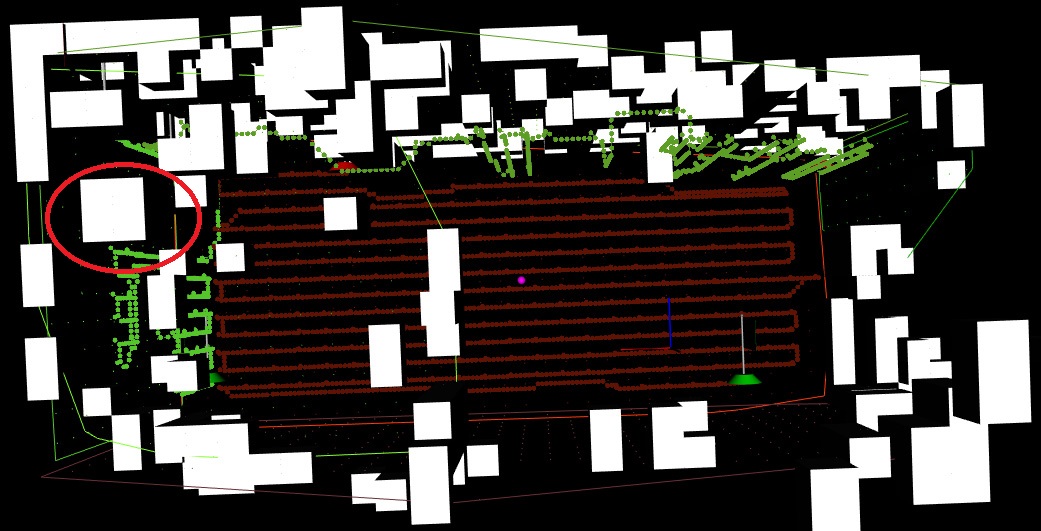}
		\caption{}
		\label{fig:office_d}
	\end{subfigure}
	\begin{subfigure}{0.5\textwidth}
		\centering
		\includegraphics[width=170 pt]{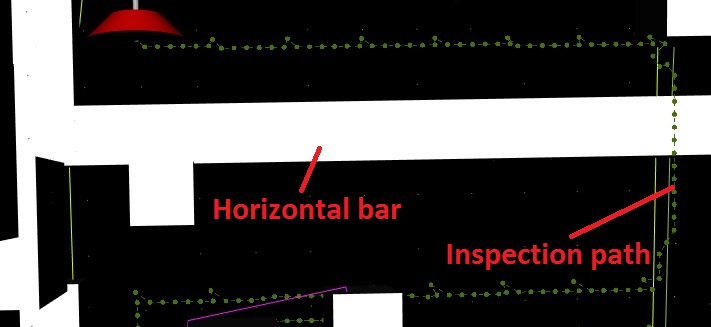}
		\caption{}
		\label{fig:office_e}
	\end{subfigure}
	\begin{subfigure}{0.5\textwidth}
		\centering
		\includegraphics[width=170 pt]{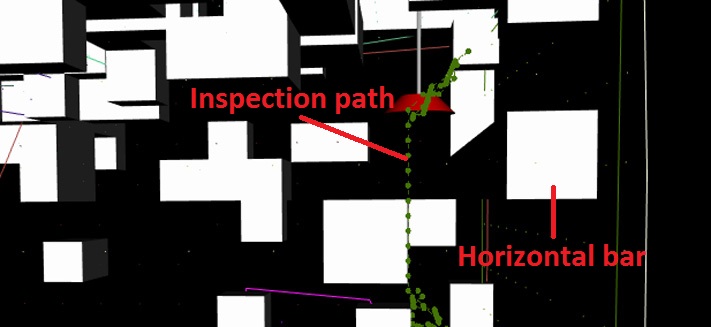}
		\caption{}
		\label{fig:office_f}
	\end{subfigure}		
	
	\centering
	\caption{Experiment with the dataset recording one floor of an office building: (a) Raw data recorded by laser scanners ; (b) Detected planar surfaces and their boundaries; (c) Path planning to inspect the surfaces of ceiling, left wall, and back wall; (d) Inspection path with the appearance of obstacles; (e) Part of inspection path avoiding an obstacle (front view); (f) Part of inspection path avoiding an obstacle (side view);}
	\label{fig:building}
\end{figure}

\begin{figure}[h!]
	\begin{subfigure}{0.5\textwidth}
		\centering
		\includegraphics[width=170 pt]{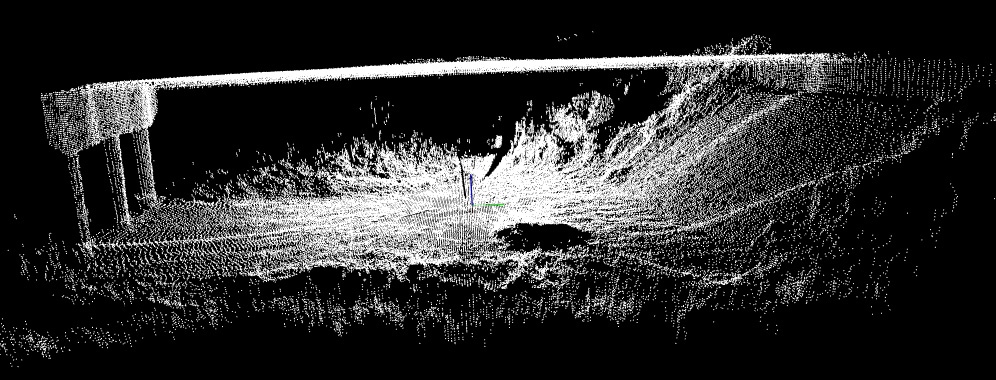}
		\caption{}
		\label{fig:bridge_a}
	\end{subfigure}%
	\begin{subfigure}{0.5\textwidth}
		\centering
		\includegraphics[width=170 pt]{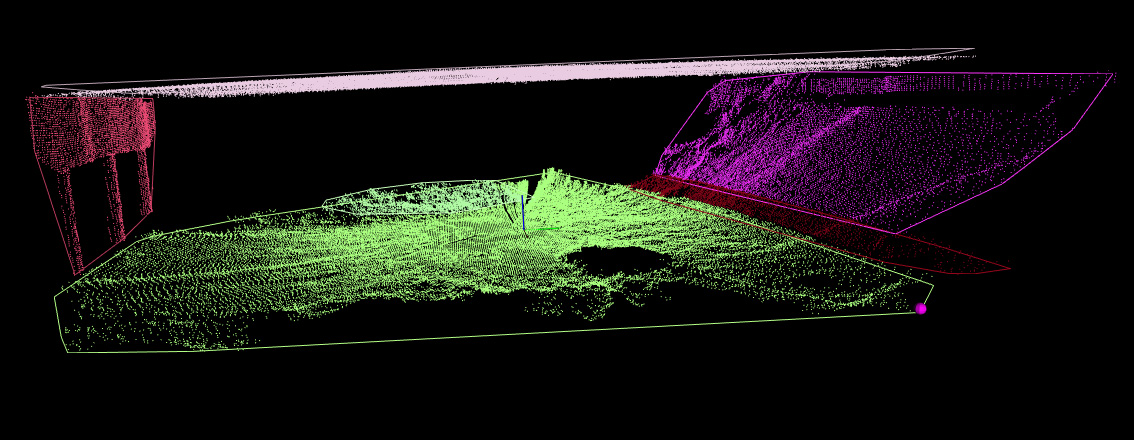}
		\caption{}
		\label{fig:bridge_b}
	\end{subfigure}
	\begin{subfigure}{0.5\textwidth}
		\centering
		\includegraphics[width=177 pt]{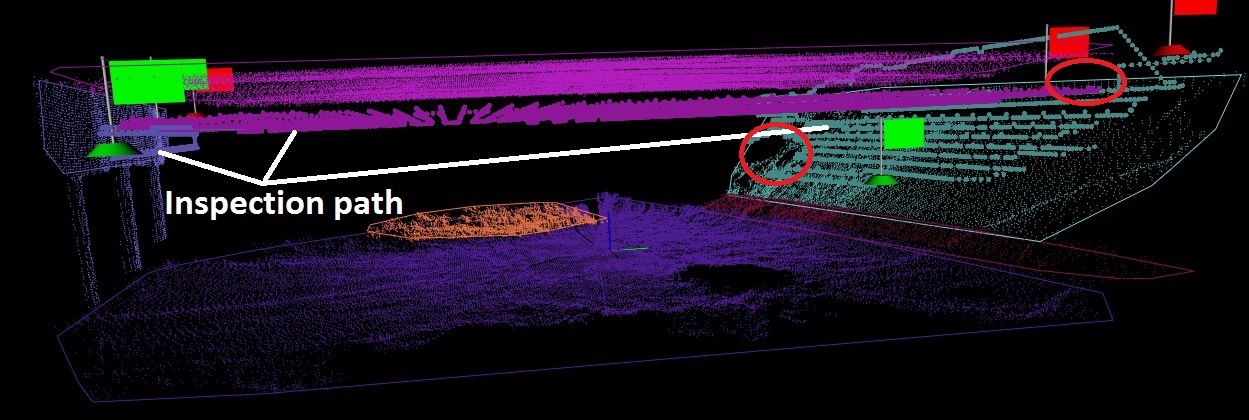}
		\caption{}
		\label{fig:bridge_c}
	\end{subfigure}%
	\begin{subfigure}{0.5\textwidth}
		\centering
		\includegraphics[width=160 pt]{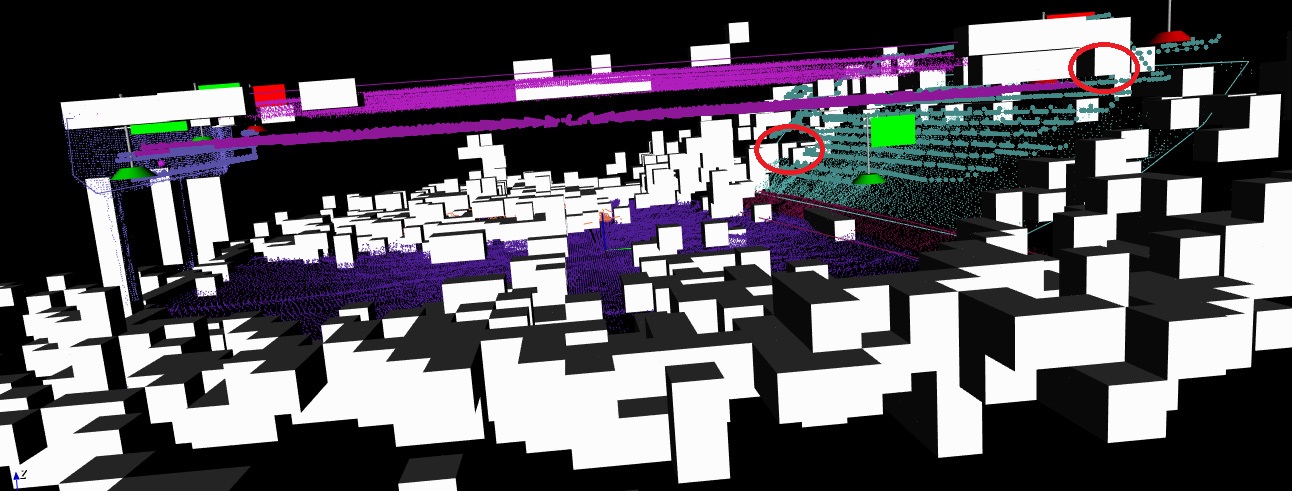}
		\caption{}
		\label{fig:bridge_d}
	\end{subfigure}
	\begin{subfigure}{0.5\textwidth}
	\centering
	\includegraphics[width=170 pt]{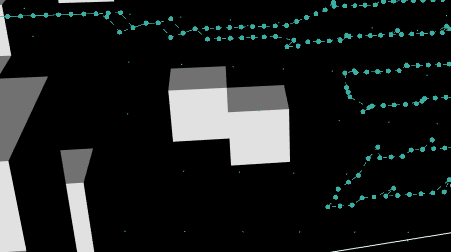}
	\caption{}
	\label{fig:bridge_e}
\end{subfigure}
	\begin{subfigure}{0.5\textwidth}
	\centering
	\includegraphics[width=160 pt]{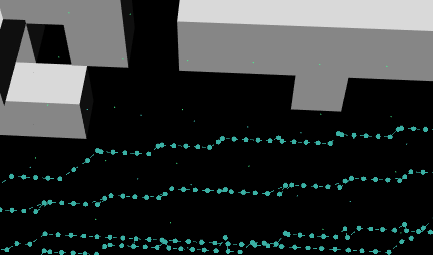}
	\caption{}
	\label{fig:bridge_f}
\end{subfigure}

	\centering
	\caption{Experiment with the dataset recording a part of a bridge: (a) Raw data recorded by laser scanners ; (b) Detected planar surfaces and their boundaries; (c) Path planning to inspect the piers, top surface, and slope surface; (d) Inspection path with the appearance of obstacles; (e) Part of inspection path avoiding obstacles at surface bottom; (f) Part of inspection path avoiding obstacles at surface top;}
	\label{fig:bridge}
\end{figure}

Experiments have been carried out on two real datasets recorded by laser scanners mounted on a UAV for inspection of an office building and a concrete bridge. The first dataset represents a floor of the building with a size of 25 m $\times$ 12 m $\times$ 8 m. The second dataset represents a part of the bridge including piers and surfaces with a size of 22 m $\times$ 10 m $\times$ 4.5 m. Figures \ref{fig:office_a} and \ref{fig:bridge_a} show the datasets in point cloud representation. In order to apply the IPP algorithm to the datasets, planar surfaces and boundaries need to extracted from them. For this task, we have developed a software for automatic interpretation of unordered point cloud data described in details in \cite{Phung2016}. The software uses the Random sample consensus (RANSAC) algorithm combined with data obtained from an inertial measurement unit (IMU) to detect planar surfaces. The convex hull algorithm is then employed to determine their boundaries. The remaining point cloud is clusterized to obstacle objects by finding the nearest neighbour in a 3D Kd-tree structure. Through the software, users are able to select the surfaces they want to inspect, as shown in Fig.~\ref{fig:office_b} and Fig.~\ref{fig:bridge_b}, respectively.

In all experiments, coefficients $w = 1$, $\varphi_1 = 0.4$, $\varphi_2 = 0.4$ are chosen for the DPSO. The number of particles is set to 100. The random mutation is executed in every three generations and the edge exchange is carried out if the random mutation does not improve the result. The parallel implementation is developed based on the CUDA platform. The programs, including both serial and parallel versions, are executed in a laptop computer with Core\textsuperscript{TM}i7 CPU and GeForce\textsuperscript{$@$} GTX 960M GPU.

\begin{figure}[h!]
	\begin{subfigure}{0.5\textwidth}
		\centering
		\includegraphics[width=170 pt]{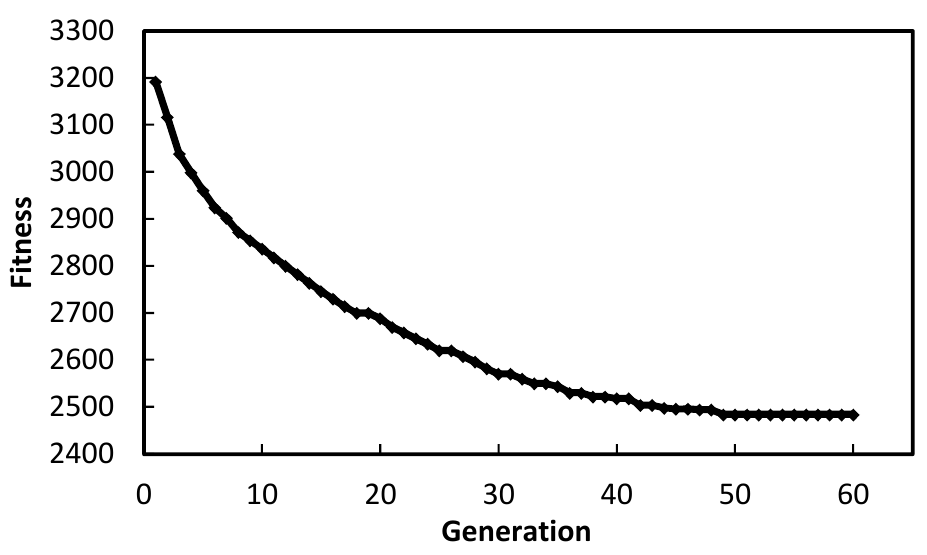}
		\caption{}
		\label{fig:convergence_a}
	\end{subfigure}%
	\begin{subfigure}{0.5\textwidth}
		\centering
		\includegraphics[width=170 pt]{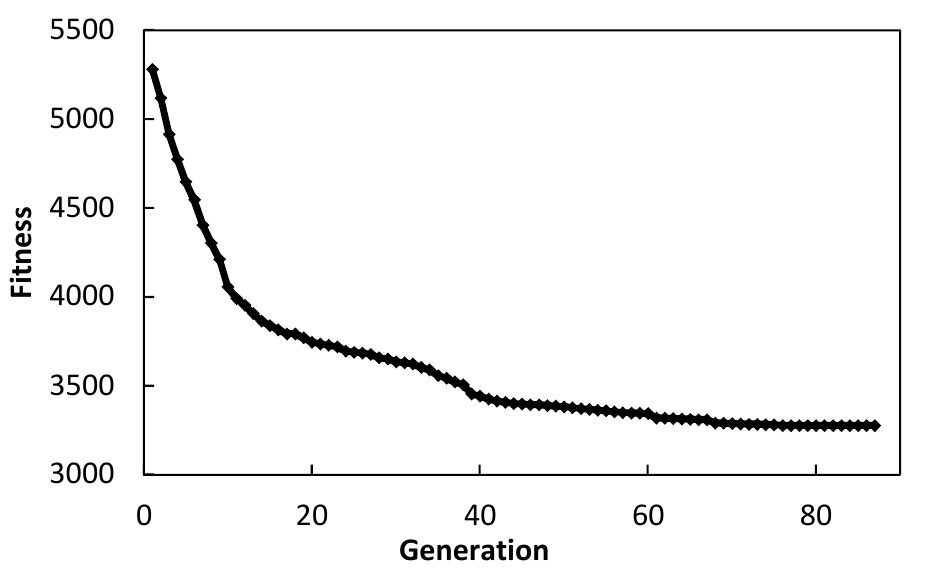}
		\caption{}
		\label{fig:convergence_b}
	\end{subfigure}
	\centering
	\caption{DPSO Convergence from datasets of: (a) Office building; (b) Concrete bridge.}
	\label{fig:convergence}
\end{figure}

\subsection{Path Generation and DPSO Convergence}
Figures~\ref{fig:office_c} and \ref{fig:bridge_c} show the paths generated to inspect three selected surfaces of each dataset. Figures~\ref{fig:office_d} and \ref{fig:bridge_d} show the paths in the appearance of obstacles. It can be seen that the back-and-forth pattern is dominant in those paths, except essential changes when having obstacles or switching between surfaces. Figures~\ref{fig:office_e} and \ref{fig:office_f} present the front and side views of a zoom-in part of the inspection path showing that obstacles were avoided. Figures~\ref{fig:bridge_e} and \ref{fig:bridge_f} show similar results for the bridge dataset. 

Figure~\ref{fig:convergence} shows the graphs of the fitness value as an objective function of the generation number for the two inspection cases of a building and a bridge. In each graph, the fitness represents the cost to traverse the inspection path. From the dataset of the office building, the DPSO by solving the extended TSP improves 22.2 \% of the travelling cost and converges within 60 generations. For the second dataset of the bridge, those numbers are 37.9 \% and 80, respectively. The difference is accounted for by the variation in size of the inspection surfaces and the structural complexity of the environments. That is to say in terms of algorithms, care should be given when considering parameters for the exploration (number of particles) and exploitation (number of generations).

\subsection{Effect of the augmentations on the DPSO}
Table~\ref{tab:effect} presents the effect of augmentations on the performance improvement over DPSO in percentage by applying our enhanced algorithm. Here, with the dataset obtained from building inspection, the deterministic initialization significantly improves the processing time by 2.8 times and slightly improves the travelling cost by 1.4 \%. Notably, the computational efficiency in terms of fast convergence actually comes from the improvement of evolving generations of the swarm by means of initialization. On the other hand, it is not surprised that the edge exchange introduces some enhancement on the travelling cost as it uses brute force transpositions. Likewise, the parallel implementation introduces the most significant impact on the computation time thanks to the parallel processing capability taking advantage of the SIMD feature of the DPSO. 

\begin{table}[h!]
	\centering
	\caption{Percent improvement of the DPSO by augmentations}
	\label{tab:effect}
	\begin{tabular}{p{3cm} p{1cm} p{2.5cm} p{1cm} p{2.5cm}} 
		\hline
		\rule{0pt}{3ex}
		\multirow{3}{*}{Algorithm} & \multicolumn{2}{c}{Building dataset} & \multicolumn{2}{c}{Bridge dataset} \\ [1ex] \cline{2-5} 
		\rule{0pt}{3ex}    
		& Time            & Travelling cost           & Time            & Travelling cost           \\ 
		& (\%)            & (\%)           & (\%)            & (\%)           		\\ [1ex]
		\hline
		\rule{-3pt}{3ex}
		Initialization    & 280  & 1.4   & 310  & 1.7  \\ [1ex]
		Random mutation   & x    & 5.0     & x    & 5.5                  \\ [1ex]
		Edge exchange     & x    & 13.8    & x    & 15.7                  \\ [1ex]
		Parallel on GPU   & 6570  &  x    & 6720  & x  \\ 
		\bottomrule
		x:\textit{ not applicable}
	\end{tabular}
\end{table}

To show consistency in the effectiveness of the proposed approach, we compare our enhanced DPSO algorithm not only with the conventional DPSO but also with an ant colony system (ACS), where the ACS is implemented as in \cite{Dorigo97}. In the comparison, each algorithm was executed over 15 trials. Table~\ref{tab:compare} shows the results expressed in the average value and the standard deviation of the processing time and the travelling cost. Compared with the ACS algorithm, our enhanced DPSO  for the bridge inspection dataset has shown on average an improvement of 15\% in the travelling cost and 87 times in the computation time. 
Owing to a significant improvement in processing time, the enhanced DPSO can be applied for real-time automated inspection.

\begin{table}[h!]
	\centering
	\caption{Comparison between the enhanced DPSO, DPSO and ACS algorithms.}
	\label{tab:compare}
	\begin{tabular}{p{2.6cm} p{1.5cm} p{2.3cm} p{1.3cm} p{2.3cm}} 
		\hline
		\rule{0pt}{3ex}
		\multirow{3}{*}{Algorithm} & \multicolumn{2}{c}{Building dataset} & \multicolumn{2}{c}{Bridge dataset} \\ [1ex] \cline{2-5} 
		\rule{0pt}{3ex}    
		& Time (s)            & ~Travelling cost           & Time (s)           & ~Travelling cost           \\       [1ex]
		\hline
		\rule{-3pt}{3ex}
		Enhanced DPSO    & 32.9$\pm$1.2  & ~~~2490.2$\pm$38.9   & 41.6$\pm$1.5  & ~~~~3358.5$\pm$59.7  \\ [2ex]
		DPSO   			 & 2253.8$\pm$25.2    & ~~~2998.1$\pm$44.3     & 2928.4$\pm$33.8    & ~~~~4130.7$\pm$65.8                  \\ [2ex]
		ACS				 & 2560.1$\pm$16.3    & ~~~2763.6$\pm$56.8    & 3617.2$\pm$23.4    & ~~~~3862.3$\pm$87.4                  \\ [1ex]
		\bottomrule
	\end{tabular}
\end{table}

\section{Conclusion}
In this paper, we have presented an enhanced discrete particle optimisation (DPSO) algorithm for solving the inspection path planning (IPP) problem that is formulated as an extended travelling salesman problem (TSP) considering simultaneously the coverage and obstacle avoidance. By augmenting with deterministic initialization, random mutation, edge exchange and parallel implementation on GPU, the proposed DPSO can greatly improve its performance in both time and travelling cost. The validity and effectiveness of the proposed technique are verified in successful experiments with two real-world datasets collected by UAV inspection of an office building and a concrete bridge. In a future work, the algorithm will be extended for inspection of non-planar surfaces and incorporation of online re-planning strategies to deal with inspection of built infrastructure of an irregular shape.

\section*{Acknowledgments}
The first author would like to acknowledge an Endeavours Research Fellowship (ERF-PDR-142403-2015) provided by the Australian Government. This work is supported by the University of Technology Sydney Data Arena Research Exhibit Grant 2016 and Vietnam National University Grant QG.16.29.

\section*{References}

\bibliography{reference}

\end{document}